\documentclass[11pt]{article}

\usepackage{acl}

\usepackage{times}
\usepackage{latexsym}
\usepackage[T1]{fontenc}
\usepackage[utf8]{inputenc}
\usepackage{microtype}

\usepackage{hyperref}
\usepackage{url}
\usepackage{booktabs}
\usepackage{amsfonts}
\usepackage{amsmath}
\usepackage{amssymb}
\usepackage{nicefrac}
\usepackage{xcolor}
\usepackage{graphicx}
\usepackage{multirow}
\usepackage{array}
\usepackage{makecell}
\usepackage[capitalise]{cleveref}
\usepackage{enumitem}
\setlist[enumerate]{itemsep=1pt,topsep=2pt,parsep=0pt,leftmargin=*}
\setlist[itemize]{itemsep=1pt,topsep=2pt,parsep=0pt,leftmargin=*}
\usepackage{placeins}      

\title{Function-Vector Heads Are Two Populations:\\Writers and Cancellers in In-Context Learning}

\author{Han-yu Wang \\
  The University of Hong Kong \\
  \texttt{henry.why@connect.hku.hk}}

\begin{document}

\maketitle
\pagestyle{plain}
\thispagestyle{plain}

\begin{abstract}
In-context learning lets a language model perform a task specified by
examples in its prompt. Function vectors capture task information in a
compact activation assembled from attention-head outputs. Across two
rule families and three Pythia models, we find two opposed functional
populations among candidate function-vector heads. \emph{Writers}
support the rule-correct answer, while \emph{cancellers}
systematically suppress it. These roles predict opposing group effects
on held-out prompts, where removing cancellers improves accuracy by
2.4 to 6.8 percentage points in all six settings. The dominant
suppressive contributions depend on task content, and the same
components can support correct predictions in another task. At the
population level, substantial support and suppression can nearly
cancel. Separating these roles explains how task-related computation
can work against correct predictions and how a small aggregate effect
can conceal strong opposing contributions.
\end{abstract}

\section{Introduction}
\label{sec:intro}

In-context learning lets a language model infer an input--output
relation from examples in its prompt while its parameters remain
fixed \citep{brown2020fewshot}. Function-vector (FV) methods isolate an
internal activation associated with the demonstrated task. For example, inserting an
antonym function vector can elicit antonyms in a prompt containing
no antonym demonstrations \citep{todd2024fv}. The extraction
procedure identifies attention heads whose task-average outputs
recover correct answers in prompts with corrupted demonstration
labels. Aggregating these outputs produces a vector that can induce
the task. Together with related work on task representations
\citep{hendel2023task,variengien2023look}, this provides a way to
localize information used in in-context learning. Induction heads
provide another account through matching and copying continuations
of earlier token sequences \citep{olsson2022induction}. Comparing
the two types across twelve models, \citet{yin2025heads} find that
FV heads have a greater effect on few-shot in-context learning
performance, especially at larger scales. These findings motivate
examining how task-sensitive heads contribute to the answer in
an individual prompt.

We find a systematic source of opposition to correct in-context
predictions within a task-sensitive head population. We select
candidate function-vector heads by their sensitivity to demonstration
labels and test their effects in two rule families and three Pythia
models. One group supports the rule-correct label while another
suppresses it. We call these
functional populations \emph{writers} and \emph{cancellers}.
Zero-ablating the canceller group increases observed accuracy by
$2.4$--$6.8$ percentage points in every setting. The suppressive
role therefore has a concrete behavioral cost. Task-sensitive
processing can work against the answer specified by the
demonstrations.

Negative heads appear in circuits for indirect object identification
\citep{wang2023ioi}. Copy suppression provides a mechanistic example
in which heads suppress context tokens that earlier components
predict too strongly, improving overall calibration
\citep{mcdougall2023copy}. These findings connect negative
contributions on individual tasks to a broader computational role.
Our finding concerns the organization of a task-related population.
Support and suppression coexist within the identified inventory,
and their balance determines its net effect under zero ablation.
For Pythia-410M on the hierarchical rule task, removing the two
groups separately produces substantial changes in opposite directions,
while removing both
has almost no net effect. The writer--canceller distinction reveals
the opposing contributions hidden by this aggregate result.

We identify the candidate inventory through label-sensitive direct
logit attribution (DLA) and validate signed head effects with paired
interventions under changes in the demonstrated rule. The validated
effects define the two roles. Membership then remains fixed during
separate and joint ablations on held-out prompts. This design
tests whether head-level assignments predict whole-group effects
and whether those effects reach the model's answer decisions.
The opposing readout shifts and accuracy changes recur across
all six model--task settings under zero ablation.

Source analysis explains how task content enters the opposing
computation. Compared with writers, cancellers devote less attention
to demonstration labels and more to the format prefix. A source's contribution
depends on both its attention weight and the value vector it supplies.
Demonstration and query content dominate these contributions for most
cancellers. For the canceller with the most negative mean DLA in each
setting, permuting source values at fixed attention removes
$52$--$83\%$ of its negative attribution. The dominant suppression thus
depends on the content combined by the head's attention pattern.

Following the same heads across tasks reveals how their roles
depend on the computation. The groups retain their opposing effects
between the two rule families. A group identified as cancellers on
one rule task supports correct antonym answers after transfer, as
does one of its heads. On country-capital prompts, that group remains suppressive while
the same head has little effect. Writers and cancellers therefore
describe how components participate in the current computation.
Together, the results establish a task-conditioned source of
opposition within the candidate inventory and explain how its
behavioral cost, content dependence, and aggregate effect fit
together.

\section{Setup}
\label{sec:setup}

The main analysis applies the same discovery and validation stages
to each pairing of two rule tasks and three Pythia models. We call
each pairing a model--task setting.

\paragraph{Models.}
The main pipeline studies three Pythia checkpoints
\citep{biderman2023pythia} ($410$M, $1$B, $1.4$B), loaded in
\texttt{fp32} with eager attention to expose per-head weights and outputs.
The checkpoints keep the architecture family fixed while varying
model size (\Cref{tab:models}).

\begin{table}[t]
\centering
\caption{Pythia checkpoints used in the six model--task settings.
$L$ is the layer count, $H$ the number of heads per layer, and
$d_{\mathrm{head}}$ the head dimension.}
\label{tab:models}
\small
\setlength{\tabcolsep}{3.5pt}
\begin{tabular}{lccccc}
\toprule
Model & arch.\ & $L$ & $H$/layer & $d_{\text{model}}$ & $d_{\text{head}}$ \\
\midrule
Pythia-$410$M    & GPT-NeoX & 24 & 16 & 1024 & 64  \\
Pythia-$1$B      & GPT-NeoX & 16 & 8  & 2048 & 256 \\
Pythia-$1.4$B    & GPT-NeoX & 24 & 16 & 2048 & 128 \\
\bottomrule
\end{tabular}
\end{table}

\paragraph{Tasks.}
\label{sec:setup:tasks}
We study two in-context rule-following tasks with binary answers.
Generating each prompt from a specified rule gives us a
correct-versus-incorrect contrast for measuring head contributions.
The two tasks share a four-shot surface form. Each prompt contains
four demonstrations followed by a query, written as
\texttt{f0,f1,Y\,\ldots\,f0q,f1q,}, where the model must emit
the label for the query pair. Feature pairs
$(f_0, f_1)$ take integer values in $\{0,\ldots,7\}^2$.
A hidden rule $z\!\in\!\{0,1,2,3\}$ determines the binary labels
$Y\!\in\!\{A,B\}$ and remains fixed within a prompt.
The model sees the labeled examples and must infer how to label
the query. The rule itself is withheld from the prompt.
For example, the rule assigning A when $f_0+f_1>7$ and B otherwise
labels the final query below as A
\begin{center}
\texttt{3,5,A 1,2,B 4,5,A 2,3,B 6,3,}\,$\to$\,\texttt{A}
\end{center}
The two rule families differ in rule structure.
In the \emph{hierarchical} family, $z$ selects one of
$\{f_0\!>\!3,\;f_1\!>\!3,\;f_0\!+\!f_1\!>\!7,\;f_0\!=\!f_1\}$.
In the \emph{modular} family, $z$ selects a modulus $m\!\in\!\{2, 3, 5, 7\}$
and assigns the binary outcome
$\mathbf{1}[(f_0\!+\!f_1)\bmod m\!=\!0]$. The shared surface form holds prompt format fixed while the rule
family changes, allowing the signed roles to be compared across two
distinct computations.

\paragraph{Readouts and effects.}
For a rule-task prompt $x$, let $y_+$ and $y_-$ denote the correct
and alternative answer labels. The readout is
\begin{equation}
R(x)=\log p(y_+\mid x)-\log p(y_-\mid x).
\label{eq:readout}
\end{equation}
This log-probability contrast measures how strongly the model
favors the correct label over its alternative. It is positive when
the correct label is more probable and zero when the two labels
are equally probable. A positive head contribution supports this
contrast. For an ablated group $G$, we write
\begin{equation}
\delta_G=\mathbb{E}_x[R^{\mathrm{abl}(G)}(x)-R(x)].
\label{eq:shift}
\end{equation}
Positive $\delta_G$ means that the correct label gains support
relative to its alternative when the group is removed. A suppressive
group is therefore expected to produce a positive ablation shift.
The held-out ablations test this prediction.

\section{Methods}
\label{sec:methods}
Causal mediation and path patching connect internal activations
to changes in model behavior
\citep{vig2020causalmediation,goldowsky2023pp}. We use paired
interventions to assign signed roles to task-sensitive candidate
heads, testing whether each head supports or suppresses the
answer under changes in the demonstrated rule.

Automated circuit discovery studies how components interact
\citep{conmy2023acdc,syed2024ap}, while sparse feature circuits
trace such interactions at the level of activation features
\citep{marks2025sfc}. Our group analysis asks how the signed roles
of individual heads combine within the selected population.
We hold the roles fixed during separate and joint ablations on
held-out prompts, then use source attribution and value interventions
to trace contributions to source content. Each main setting uses
$192$ discovery prompts, $200$ validation prompt pairs, and $500$
held-out evaluation prompts.

\subsection{Screening task-sensitive heads}
\label{sec:methods:dla}
An attention head reads information from other token positions
and adds a vector to the residual stream, the shared representation
updated by successive transformer components. The output layer
maps the final representation to token scores. Direct logit
attribution applies the answer-token directions to an individual
head's contribution, measuring how it aligns with support for the
correct versus incorrect label
\citep{elhage2021mathematical,geva2022dla}.
For head $h$, let $c_h(x)$ denote its output after the output
projection. The label-unembedding contrast $u(x)$ is the difference
between the output-layer directions for the correct and alternative
labels. Projecting $c_h(x)$ onto $u(x)$ gives a signed contribution
to the correct-label readout.

We subtract the mean attribution obtained after permuting the
demonstration labels. This centers the screen on the contribution
that depends on those labels. A $20$-seed permutation analysis with
Benjamini--Hochberg correction controls the false discovery rate
(FDR) at $q=0.10$ \citep{benjamini1995controlling}. We combine the
significant heads with the top-$K$ heads by absolute uncentered DLA. For the
hierarchical task, $K$ is $27$, $24$, and $20$ at $410$M, $1$B, and
$1.4$B, respectively. The modular task uses $K=20$ at each scale.
The resulting shortlist $\mathcal S$ includes contributions in both directions.

\subsection{Assigning roles from paired effects}
\label{sec:methods:pp}
Each prompt pair holds the query fixed while changing the
demonstrated rule. We compare the clean and rule-corrupted readouts
and the readout after a head-specific direct-path intervention.
The intervention targets the head's contribution along its direct
route to the answer \citep{wang2023ioi,goldowsky2023pp}.
All three readouts use the same label-contrast orientation within
a pair. We define the full swing and the head-specific effect as
\begin{align}
s_i &= R_i^{\mathrm{clean}}-R_i^{\mathrm{corr}},\\
e_{hi} &= R_{hi}^{\mathrm{direct}}-R_i^{\mathrm{corr}}.
\end{align}
We normalize within each pair and then average,
\begin{equation}
d_h=\frac{1}{|\mathcal I|}\sum_{i\in\mathcal I}\frac{e_{hi}}{s_i},
\qquad \mathcal I=\{i:s_i>0.5\}.
\label{eq:directpct}
\end{equation}
The ratio expresses a head's effect relative to the change
produced by changing the demonstrated rule. A positive ratio
moves the readout toward the clean-prompt outcome, and a negative
ratio moves it away. The threshold retains pairs with an oriented full swing greater
than $0.5$\,nats. The six settings retain $96$--$121$ pairs each.
Paired-prompt bootstraps use $10{,}000$ resamples.

Within each model--task setting, we apply Benjamini--Hochberg
correction at $q=0.10$ to the two-sided direct-effect tests. The
passing heads form $\mathcal V$. Among them, heads with
$d_h\geq0.05$ are \emph{writers} ($\mathcal W$), and heads with
$d_h\leq-0.05$ are \emph{cancellers} ($\mathcal C$). Their union
$\mathcal F=\mathcal W\cup\mathcal C$ is the selected signed
inventory. These two functional populations are defined for a
model, task, and readout. The subsequent analyses follow these heads across new prompts
and tasks. App.~\ref{app:head_inventory} lists the complete inventory
and paired-effect estimates.

\subsection{Testing roles through held-out ablation}
\label{sec:methods:groupcanc}
Roles are assigned from interventions on individual heads. We
then hold those assignments fixed while ablating each group and
their union on $500$ held-out prompts. The remaining network can
respond to a group intervention through compensatory changes
\citep{mcgrath2023hydra,rushing2024selfrepair}. Zero ablation
removes the selected outputs. Mean ablation replaces them with
their task means.
The separate interventions test whether the assigned roles predict
the direction of held-out effects and measure each group's effect
size. The union intervention measures how the two groups combine
at the readout.

We report mean shifts $\delta_{\mathcal W}$,
$\delta_{\mathcal C}$, and $\delta_{\mathcal F}$ with paired-prompt
bootstrap $95\%$ CIs.
Evaluation stability is assessed over five additional random
splits, each containing $500$ prompts in the evaluation half,
while keeping the original role assignments fixed.

\subsection{Tracing contributions to source content}
\label{sec:methods:qk}
We trace each head's contribution to the material it reads from
the prompt. We group attention from the final prompt position by
source, separating the beginning-of-sequence token (BOS), the format
prefix (a comma followed by a space), demonstration inputs,
demonstration labels, and query input. The means over
$32$ prompts describe how each group distributes attention.

To measure each source's contribution, we decompose the head output as
\begin{equation}
c_h(x)=\sum_j\alpha_{hj}(x)W_{O,h}v_{hj}(x),
\label{eq:source}
\end{equation}
where $\alpha_{hj}$ is the attention weight, $v_{hj}$ the value
vector at source position $j$, and $W_{O,h}$ the projection into
the residual stream. Projecting the source terms onto the label
contrast gives their signed contributions. We call a canceller's
attribution content-driven when the summed attribution from
demonstration and query positions has a larger magnitude than
the summed attribution from BOS and format positions.

This decomposition separates the amount of attention allocated to a
position from the contribution of its value vector. A position with
relatively little attention can still make a substantial contribution
if its transformed value projects strongly onto the label contrast.
The attention profile and source attribution therefore show where
the head looks and how the retrieved material contributes to the answer.

To test the dependence on source assignment, we hold attention
weights fixed, permute value vectors across source positions,
and recompute the head-output projection. This intervention
changes which values are combined with each attention weight.
We apply it to the canceller with the most negative mean DLA in
each setting and examine Pythia-410M's L11.H4 in detail. This notation
denotes head $4$ in layer $11$, using zero-based indices.

\subsection{Following roles across tasks and selection rules}
\label{sec:methods:specificity}
We first transfer the signed groups between the two rule families
at each Pythia scale. For Pythia-410M, we also apply the groups to
four-shot antonym and country-capital prompts in a
\texttt{word:answer} format. Vocabulary-task effects measure the
change in the correct-answer logit after zero ablation.
This follows the same source-task groups under a different task
and answer readout.

To examine the effect of selection by magnitude, we construct a
separate comparator that ranks heads by the absolute mean-ablation
effect on the current task. We denote its top-$K$ set by MA-$K$
and measure its coverage of writers and cancellers in $\mathcal F$.

\begin{figure*}[t]
\centering
\includegraphics[width=\linewidth]{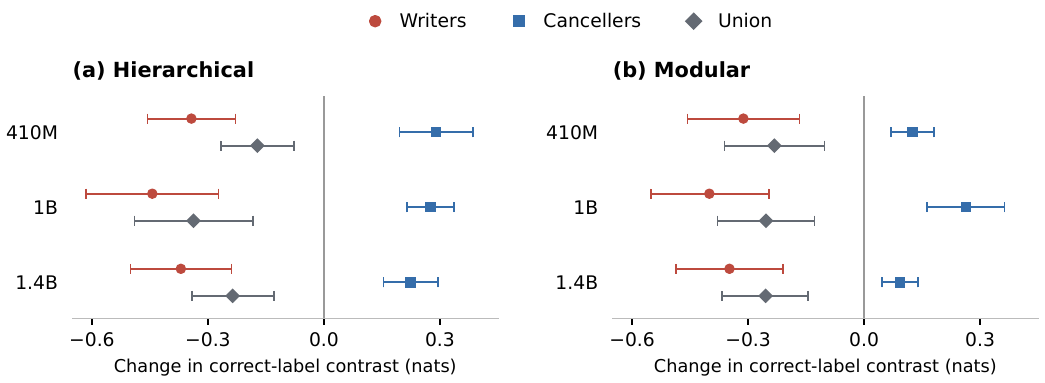}
\caption{\small\textbf{Held-out group-ablation effects.}
Mean-ablation readout shifts for writers, cancellers, and their union
in the six main model--task settings. Writer ablation lowers the
correct-label contrast and canceller ablation raises it in every
setting. Each estimate uses $500$ prompts. Error bars show paired-prompt
bootstrap $95\%$ CIs.}
\label{fig:headline}
\end{figure*}

\section{Results}
\label{sec:results}

DLA screening yields $20$--$30$ candidates per setting.
Paired validation retains $12$--$23$ significant heads, of which
$11$--$15$ meet the signed effect-size threshold. This gives $54$
writer and $27$ canceller assignments across the six settings.
These counts refer to head--task assignments, since a head can
appear in both tasks at the same model size.

\subsection{Cancellers oppose correct answers on held-out prompts}
\label{sec:results:groupcanc}
Roles assigned on the validation prompts predict the direction of
group effects on held-out prompts. Zero-ablating cancellers raises
the correct-label readout by $+0.13$ to $+0.29$\,nats in every main
setting. Writer ablation lowers it by $0.25$ to $0.99$\,nats
(\Cref{tab:groupcanc}). Mean ablation gives the same directions
across the six settings (\Cref{fig:headline}), with canceller shifts
of $+0.09$ to $+0.29$\,nats (App.~\ref{app:groupcanc}).
The same assigned groups also produce opposing effects across
all five evaluation splits in every main setting
(App.~\ref{app:split_detail}).

Under zero ablation, the joint effect differs from the sum of
the two separate effects by at most $0.09$\,nats in each setting.
This near-additivity explains how large contributions in opposite
directions can combine into a small net readout change.

Hierarchical-410M shows near-complete cancellation. Removing its writers
changes the readout by $-0.25$\,nats, and removing its cancellers changes
it by $+0.29$\,nats. Removing both produces a shift close to zero.
Modular-1B shows a different balance. Its writer ablation shifts the
readout by $-0.99$\,nats, canceller ablation by $+0.21$\,nats, and
joint ablation by $-0.78$\,nats. Removing the writers therefore
accounts for most of the joint decrease, while removing the
cancellers offsets part of it.

\begin{table}[t]
\centering
\caption{\textbf{Zero-ablation effects.} Readout shifts for writers
($\mathcal W$), cancellers ($\mathcal C$), and their union ($\mathcal F$).
$|\mathcal V|$ counts heads passing paired-validation FDR. Writer
and canceller counts apply the signed effect-size thresholds in
\S\ref{sec:methods:pp}. Mean-ablation results appear in
App.~\ref{app:groupcanc}.}
\label{tab:groupcanc}
\small
\setlength{\tabcolsep}{4.5pt}
\begin{tabular}{lrrrrrr}
\toprule
Setting & $|\mathcal{V}|$ & $|\mathcal{W}|$ & $|\mathcal{C}|$ & $\delta_{\mathcal{W}}$ & $\delta_{\mathcal{C}}$ & $\delta_{\mathcal{F}}$ \\
\midrule
hier-410M & 22 &  8 & 6 & $-0.25$ & $+0.29$ & $+0.00$ \\
hier-1B   & 23 &  9 & 4 & $-0.56$ & $+0.16$ & $-0.43$ \\
hier-1.4B & 19 & 10 & 4 & $-0.32$ & $+0.15$ & $-0.13$ \\
mod-410M  & 12 &  8 & 3 & $-0.39$ & $+0.16$ & $-0.14$ \\
mod-1B    & 16 &  8 & 6 & $-0.99$ & $+0.21$ & $-0.78$ \\
mod-1.4B  & 19 & 11 & 4 & $-0.49$ & $+0.13$ & $-0.38$ \\
\bottomrule
\end{tabular}
\end{table}

\begin{figure*}[t]
\centering
\includegraphics[width=\linewidth]{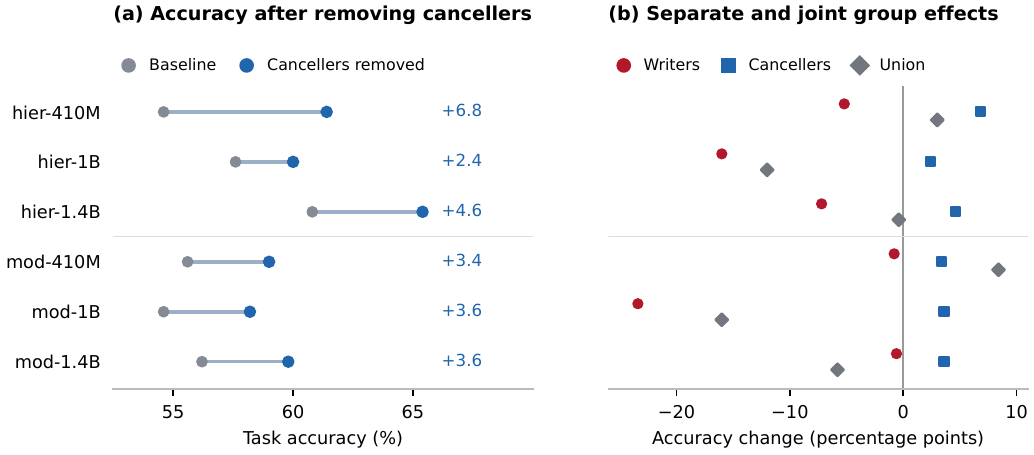}
\caption{\textbf{Behavioral effects of group ablation.}
(a) Observed accuracy before and after zero-ablating cancellers, with
gains in percentage points. (b) Accuracy changes after separately
ablating writers, cancellers, and their union. Each estimate uses the
same $500$ held-out prompts within its setting.}
\label{fig:intervention}
\end{figure*}

\paragraph{Removing cancellers improves observed accuracy.}
\label{sec:results:intervention}
On the same $500$ held-out prompts per setting, zero-ablating
cancellers increases observed accuracy in all six settings
(\Cref{fig:intervention}). The gains range from $2.4$ to $6.8$
percentage points. Hierarchical-410M increases from $54.6\%$ to
$61.4\%$, corresponding to $34$ additional correct answers in
aggregate. Hierarchical-1.4B increases from $60.8\%$ to $65.4\%$.
App.~\ref{app:accuracy} reports the complete before-and-after values.

Writer ablation decreases accuracy in every setting, by $0.6$ to
$23.4$ percentage points. Joint ablation gives a different pattern,
with increases in two settings and decreases in four.

\paragraph{Readout effects and answer decisions.}
For modular-410M, joint zero ablation decreases the mean readout by
$0.14$\,nats while increasing accuracy by $8.4$ percentage points.
The mean readout shift reflects both the direction and size of changes
across prompts. Accuracy counts correct answers, so a small shift that
changes an answer can improve accuracy while a larger shift on another
prompt leaves its answer unchanged and drives the mean in the opposite
direction.

Replacing cancellers with their task-mean outputs changes accuracy
by $-0.2$ to $+7.4$ percentage points. Writer replacement lowers
accuracy on the hierarchical task and raises it on the modular task
(App.~\ref{app:accuracy}), while its mean readout shifts remain
negative throughout.

\subsection{Dominant suppression depends on source content}
\label{sec:results:qk}
\begin{figure*}[!tb]
\centering
\includegraphics[width=\linewidth]{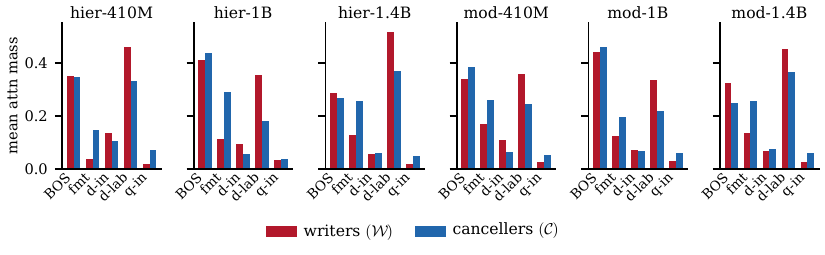}
\caption{\small\textbf{Writer and canceller attention sources.}
Mean per-head attention from the final prompt position for writers
(red) and cancellers (blue), grouped by source bucket. Cancellers allocate
less mass to demonstration labels and more to the format prefix in all
six main settings. BOS denotes the beginning-of-sequence token, fmt the format prefix,
d-in demonstration inputs, d-lab demonstration labels, and q-in the
query input.}
\label{fig:qksource}
\end{figure*}

The suppressive population affects both answer support and task
accuracy. We next trace the negative contributions to the source
content processed by its heads. Grouping attention by
source location reveals two consistent
differences across the six main settings (\Cref{fig:qksource}).
Writers place $34$--$52\%$ of their attention on demonstration
labels. Cancellers allocate $9$--$17$ percentage points less
attention there and more to the format prefix. The prefix receives
$15$--$29\%$ of canceller attention and $4$--$17\%$ of writer
attention.

Source attribution then measures what the attended material
contributes to the answer. Despite attending less to demonstration
labels, $20$ of the $27$ canceller assignments have source attribution
dominated by demonstration and query content. In each setting,
the canceller with the most negative mean DLA has content-driven
attribution. Its mean DLA has a magnitude of $0.14$ to $0.22$\,nats,
compared with less than $0.10$\,nats for each BOS- or format-dominated
canceller
(App.~\ref{app:contentvssink}).

\begin{figure*}[t]
\centering
\includegraphics[width=\linewidth]{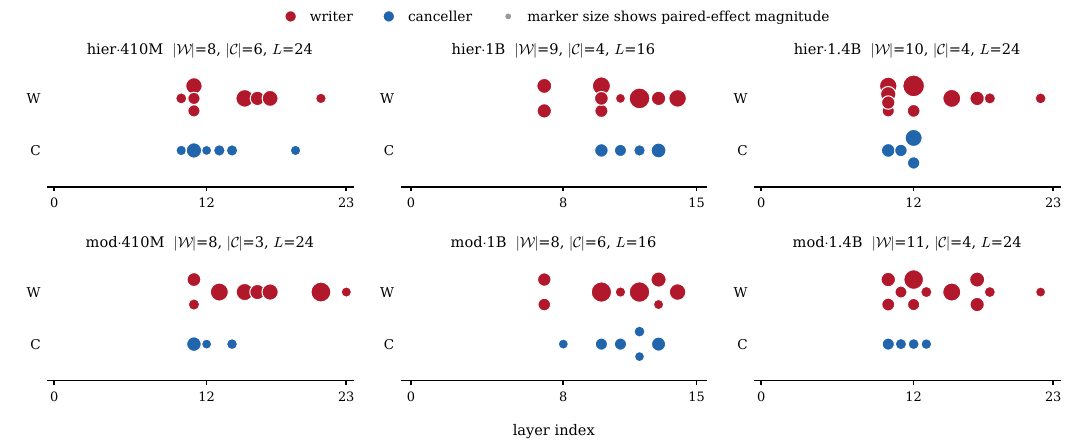}
\caption{\textbf{Layer distribution of writers and
cancellers.} Writers (red) and cancellers (blue) by layer, one
panel per setting. Marker area increases with the magnitude of the head's paired
effect $|d_h|$. Cancellers cluster in early-to-mid layers and
$23$ of $27$ canceller assignments have a writer in an earlier layer in the same setting, while later
writers continue beyond the last canceller.}
\label{fig:layergeom}
\end{figure*}

\paragraph{From source attribution to value dependence.}
\label{sec:results:casestudy}
L11.H4 has the most negative mean DLA among cancellers in both
Pythia-410M rule tasks.
Almost all of its negative DLA comes from demonstration-label
positions. We therefore examine how its contribution depends on
the values read from those positions.

\begin{figure*}[t]
\centering
\includegraphics[width=\linewidth]{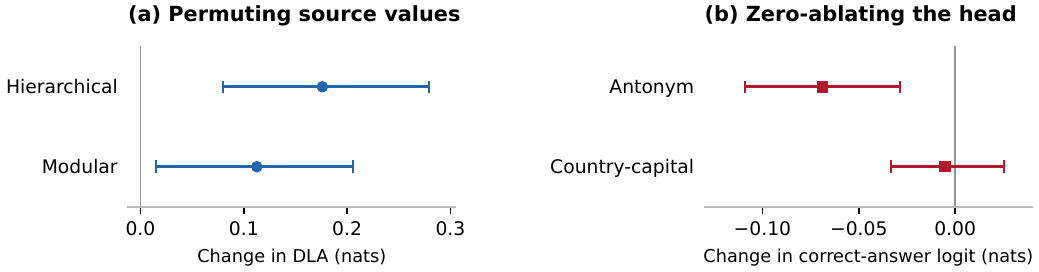}
\caption{\textbf{Content dependence and task-dependent effects in L11.H4.}
(a) Permuting source values at fixed attention reduces the negative
DLA on both rule tasks. (b) Zero-ablating the same head on vocabulary
tasks lowers the correct antonym logit and has a small effect on
country-capital answers.
Error bars show paired-prompt bootstrap $95\%$ CIs, using $200$ prompts
per rule task and $100$ per vocabulary task.}
\label{fig:casestudy}
\end{figure*}

Permuting value vectors across source positions changes the DLA by
$+0.176$ on the hierarchical task and $+0.113$ on the modular task,
removing $82\%$ and $52\%$ of the baseline negative contribution
(\Cref{fig:casestudy}). Both changes have bootstrap CIs excluding
zero. The intervention disrupts the correspondence between attention
weights and source values. Together with the source decomposition,
it connects L11.H4's suppressive contribution to the demonstration
content that the head reads.

This dependence also appears beyond L11.H4. Source-value shuffling
removes $52$--$83\%$ of the dominant canceller's negative DLA in
each main setting (\Cref{tab:valueshuffle}). The recurring effect
shows that the suppressive attribution depends on which source
values receive each attention weight.

\paragraph{Layer distribution.}
Writers and cancellers occupy overlapping layer ranges.
Cancellers cluster in early-to-middle layers, and $23$ of the $27$
assignments have a writer earlier in the same model--task setting.
In the $24$-layer models, writers extend to layers $21$--$23$, beyond
the deepest canceller at layer $19$ (\Cref{fig:layergeom}). 

\subsection{The same components change roles across tasks}
\label{sec:results:robustness}
The writer and canceller groups retain their opposing effects across
all six ordered transfers between the rule families. Both group-effect
CIs exclude zero (App.~\ref{app:rule_transfer_detail}). These transfers
keep the prompt format and answer labels fixed while changing the
demonstrated rule family.

\paragraph{Transfer to vocabulary tasks.}
\label{sec:results:vocabtransfer}
Vocabulary tasks change both the demonstrated operation and the
answer readout. We apply the Pythia-410M source-task groups to
four-shot antonym and country-capital prompts. Writer-group
ablation lowers the correct-answer logit in all four source--target
comparisons. Canceller-group ablation raises it on both
country-capital targets, by $+0.14$ and $+0.23$\,nats.
On antonyms, ablating the modular-source cancellers produces a small
shift of $+0.03$ whose CI includes zero. Ablating the hierarchical-source
cancellers lowers the correct-answer logit by $1.56$\,nats, showing
that this group supports antonym predictions (\Cref{tab:vocabtransfer}).

L11.H4 shows a corresponding change at single-head resolution.
Its ablation lowers the correct-answer logit on antonyms by
$0.069$\,nats (CI $[-0.109,-0.028]$), while the country-capital
effect is close to zero (\Cref{fig:casestudy}b). A head's signed
role therefore describes its contribution to the particular task
and readout under study.

The country-capital results show that group and individual-head
roles can change differently. Both source-task canceller groups
retain a suppressive effect, while L11.H4's individual effect becomes
small. On antonyms, the head and the hierarchical-source group both
support the answer. Thus, following the members of a transferred
group reveals changes that its aggregate effect can leave unresolved.

\subsection{Magnitude ranking recovers different functional roles}
\label{sec:results:fvoverlap}
A large absolute ablation effect can reflect either support or
suppression. We therefore use the signed inventory to interpret which
functional roles magnitude ranking recovers. Across the hierarchical
settings, MA-$20$ recovers $9/14$ cancellers ($64\%$) and $1/27$
writers ($4\%$). Across the modular settings, it recovers $16/27$
writers ($59\%$) and $1/13$ cancellers ($8\%$)
(App.~\ref{app:fv_overlap}). Among the validated heads, the same
magnitude criterion therefore favors opposite functional roles
across the two rule families.

\section{Discussion}
\label{sec:discussion}

Task-sensitive head computation can systematically suppress the
correct in-context answer. The canceller groups retain that role
on new prompts, and removing them improves observed accuracy
in all six model--task settings. The writer--canceller distinction thus
connects opposition within a task-related population to the
model's success in following the demonstrated rule.

This opposition also changes how the inventory's aggregate effect
should be understood. A small joint-ablation effect can arise
from substantial contributions in both directions. Measuring the
groups separately reveals how much support and suppression enter
the net result.
Self-repair studies show how downstream activity can compensate
for an ablated component, including through changes in suppression
\citep{mcgrath2023hydra,mcdougall2023copy,rushing2024selfrepair}.
The group effects here include the contributions of the remaining
network after each intervention. Their near-additivity under zero
ablation shows that separating the two roles gives a useful account
of the net intervention effect across the six settings.

The source analyses place dominant suppression within the
processing of task content. Attention allocation describes where
heads read, while value projections determine how the retrieved
material contributes to the answer. Their combination explains
why greater attention to format tokens can coexist with
content-dominated attribution. L11.H4 makes the relationship
concrete. Almost all of its negative projection comes from
demonstration-label positions, and changing the assignment of
values to attention weights sharply reduces that projection.
The corresponding interventions on dominant cancellers in the
other settings show the same dependence on source content.

Transfer shows how the same components participate constructively
in one task and oppose the answer in another. The changed roles
of L11.H4 and the hierarchical-source group on antonyms make this
visible at both individual and group levels. Population membership
traces which components are reused, while signed effects reveal
what their reuse accomplishes in the new computation.

For function-vector research, this raises a specific question
about how task-inducing activations relate to heads that support
or oppose individual answers. Task-vector experiments establish
that an aggregated activation can induce a task in a new context
\citep{todd2024fv,hendel2023task}. Our results suggest that such
task-level utility can coexist with functional opposition among
the computations associated with the task. Connecting the two
levels would explain how a useful representation is implemented
by a population whose members make different contributions to
task execution. The writer--canceller distinction makes those
contributions available for direct analysis.

\section{Conclusion}
\label{sec:conclusion}

Candidate function-vector heads contain two opposed functional
populations whose combined effects shape in-context predictions.
Writers support the rule-correct answer, and cancellers impose
a systematic opposing influence. Removing cancellers improves
observed accuracy across all six rule settings. Their dominant
negative projections depend on source content, and the same
components can assume supportive roles on another task.
Separating these task-dependent roles explains both the behavioral
cost of suppression and how substantial contributions can be
concealed by a small aggregate effect.

\section*{Limitations}
The main analysis examines two synthetic rule families and three
Pythia models. Their shared format supports controlled comparisons
of signed contributions. Vocabulary transfer changes the task and
answer readout together. Further controlled transfers could vary
these factors separately to trace which changes in the computation
produce a change in role.

Source attribution and value shuffling establish content dependence
at the head-output level. Following the relevant activations through
subsequent components would connect this local dependence to a more
complete circuit account.

Our signed inventory and MA-$K$ comparator are defined by the
interventions in this study. Canonical FV extraction selects heads
by signed average indirect effects in corrupted contexts
\citep{todd2024fv}. Applying both selection procedures under matched
conditions would establish how the resulting inventories overlap.

\bibliography{refs}

\clearpage
\appendix
\renewcommand{\thesection}{\Alph{section}}
\onecolumn

\section{Experimental settings and validated heads}
\label{app:repro}
\label{app:head_inventory}
The main models are the final Pythia checkpoints
\texttt{EleutherAI/pythia-410m}, \texttt{pythia-1b}, and
\texttt{pythia-1.4b}, using \texttt{fp32} and eager attention.
Discovery uses prompt seeds $42$ and $43$. Paired validation uses
pair seeds $42$ and $142$. At $410$M, $1$B, and $1.4$B, respectively,
the hierarchical settings retain $116$, $112$, and $121$ of $200$
prompt pairs, and the modular settings retain $96$, $96$, and $98$.
Main paired-effect and group-effect intervals use $10{,}000$ bootstrap
draws. Vocabulary-transfer intervals use $2{,}000$ draws.

Table~\ref{tab:head_inventory_all} lists the $81$ head--task assignments
selected by the significance and effect-size criteria in
\S\ref{sec:methods:pp}. Head indices are zero-based. W and C denote
writers and cancellers. The data and supplementary measurements are
available at \url{https://github.com/henryhyw/function-vectors-two-populations}.

\begin{table}[!ht]
\centering
\footnotesize
\setlength{\tabcolsep}{5pt}
\renewcommand{\arraystretch}{0.97}
\caption{\textbf{Complete signed head inventory.} Paired effects and per-head bootstrap intervals are expressed as percentages of the oriented readout swing in \Cref{eq:directpct}.}
\label{tab:head_inventory_all}
\begin{tabular}{lcrl@{\hspace{20pt}}lcrl}
\toprule
\multicolumn{4}{c}{Hierarchical} & \multicolumn{4}{c}{Modular} \\
\cmidrule(lr){1-4}\cmidrule(lr){5-8}
Head & Role & $100d_h$ & $95\%$ CI & Head & Role & $100d_h$ & $95\%$ CI \\
\midrule
\multicolumn{8}{c}{\textbf{Pythia-410M}} \\
L10.H9 & W & $+8.0$ & $[+5.1,+11.0]$ & L11.H2 & W & $+16.5$ & $[+9.3,+22.1]$ \\
L11.H1 & W & $+11.1$ & $[+8.1,+14.2]$ & L11.H14 & W & $+8.6$ & $[+4.0,+12.5]$ \\
L11.H2 & W & $+26.3$ & $[+20.8,+31.8]$ & L13.H5 & W & $+33.1$ & $[+21.6,+45.4]$ \\
L11.H14 & W & $+11.3$ & $[+8.3,+14.5]$ & L15.H5 & W & $+29.3$ & $[+25.1,+33.8]$ \\
L15.H5 & W & $+31.4$ & $[+27.6,+35.2]$ & L16.H13 & W & $+21.3$ & $[+18.4,+24.3]$ \\
L16.H13 & W & $+18.3$ & $[+15.1,+21.4]$ & L17.H11 & W & $+25.6$ & $[+20.2,+30.5]$ \\
L17.H11 & W & $+24.0$ & $[+18.6,+29.3]$ & L21.H5 & W & $+41.9$ & $[+34.8,+48.6]$ \\
L21.H1 & W & $+6.4$ & $[+3.5,+9.3]$ & L23.H2 & W & $+5.7$ & $[+4.9,+6.5]$ \\
L10.H8 & C & $-6.7$ & $[-9.9,-3.7]$ & L11.H4 & C & $-19.0$ & $[-33.3,-3.3]$ \\
L11.H4 & C & $-22.1$ & $[-36.7,-8.3]$ & L12.H0 & C & $-5.2$ & $[-7.1,-3.3]$ \\
L12.H0 & C & $-5.4$ & $[-7.4,-3.5]$ & L14.H3 & C & $-7.3$ & $[-9.1,-5.6]$ \\
L13.H1 & C & $-9.3$ & $[-15.6,-3.0]$ &  & & &  \\
L14.H3 & C & $-9.0$ & $[-10.5,-7.5]$ &  & & &  \\
L19.H15 & C & $-6.2$ & $[-7.2,-5.3]$ &  & & &  \\
\midrule
\multicolumn{8}{c}{\textbf{Pythia-1B}} \\
L7.H0 & W & $+20.4$ & $[+16.5,+24.7]$ & L7.H0 & W & $+15.2$ & $[+9.7,+21.2]$ \\
L7.H2 & W & $+18.3$ & $[+14.5,+22.5]$ & L7.H2 & W & $+12.5$ & $[+8.1,+17.6]$ \\
L10.H4 & W & $+33.3$ & $[+23.7,+43.1]$ & L10.H4 & W & $+44.0$ & $[+33.0,+56.0]$ \\
L10.H5 & W & $+14.1$ & $[+12.0,+16.4]$ & L11.H2 & W & $+6.3$ & $[+4.7,+8.1]$ \\
L10.H6 & W & $+16.2$ & $[+11.6,+21.3]$ & L12.H0 & W & $+44.8$ & $[+37.0,+53.5]$ \\
L11.H2 & W & $+5.2$ & $[+3.7,+6.8]$ & L13.H2 & W & $+5.7$ & $[+3.9,+7.9]$ \\
L12.H0 & W & $+46.1$ & $[+38.2,+54.4]$ & L13.H5 & W & $+21.3$ & $[+16.3,+27.0]$ \\
L13.H5 & W & $+18.3$ & $[+14.3,+22.5]$ & L14.H2 & W & $+26.6$ & $[+21.2,+31.6]$ \\
L14.H2 & W & $+31.2$ & $[+27.2,+35.0]$ & L8.H7 & C & $-5.4$ & $[-9.8,-1.2]$ \\
L10.H2 & C & $-15.4$ & $[-18.6,-12.5]$ & L10.H2 & C & $-11.1$ & $[-14.5,-8.0]$ \\
L11.H6 & C & $-10.9$ & $[-15.9,-5.9]$ & L11.H6 & C & $-11.2$ & $[-16.0,-6.5]$ \\
L12.H5 & C & $-9.3$ & $[-11.5,-7.3]$ & L12.H1 & C & $-5.2$ & $[-8.9,-1.7]$ \\
L13.H1 & C & $-20.2$ & $[-25.3,-15.5]$ & L12.H4 & C & $-7.7$ & $[-12.6,-3.2]$ \\
 & & &  & L13.H1 & C & $-17.5$ & $[-22.6,-13.2]$ \\
\midrule
\multicolumn{8}{c}{\textbf{Pythia-1.4B}} \\
L10.H0 & W & $+30.7$ & $[+24.6,+37.1]$ & L10.H0 & W & $+18.6$ & $[+13.7,+23.5]$ \\
L10.H8 & W & $+11.1$ & $[+9.1,+13.2]$ & L10.H11 & W & $+13.4$ & $[+10.8,+16.4]$ \\
L10.H9 & W & $+22.1$ & $[+17.6,+27.1]$ & L11.H1 & W & $+10.5$ & $[+6.9,+14.5]$ \\
L10.H11 & W & $+14.8$ & $[+12.1,+17.6]$ & L12.H1 & W & $+11.1$ & $[+8.3,+13.7]$ \\
L12.H1 & W & $+13.0$ & $[+10.5,+15.5]$ & L12.H15 & W & $+40.9$ & $[+31.8,+51.0]$ \\
L12.H15 & W & $+51.1$ & $[+40.5,+62.2]$ & L13.H6 & W & $+8.4$ & $[+6.7,+10.3]$ \\
L15.H11 & W & $+32.6$ & $[+29.2,+35.9]$ & L15.H11 & W & $+33.3$ & $[+30.7,+35.8]$ \\
L17.H7 & W & $+18.7$ & $[+15.2,+22.0]$ & L17.H7 & W & $+20.8$ & $[+16.6,+24.6]$ \\
L18.H0 & W & $+8.7$ & $[+6.5,+11.0]$ & L17.H10 & W & $+18.0$ & $[+16.2,+19.7]$ \\
L22.H2 & W & $+7.6$ & $[+5.1,+9.8]$ & L18.H0 & W & $+7.9$ & $[+5.6,+10.1]$ \\
L10.H7 & C & $-16.2$ & $[-24.1,-8.4]$ & L22.H2 & W & $+5.2$ & $[+2.6,+7.7]$ \\
L11.H2 & C & $-12.2$ & $[-14.2,-10.3]$ & L10.H7 & C & $-11.0$ & $[-18.9,-3.9]$ \\
L12.H0 & C & $-11.8$ & $[-18.4,-5.3]$ & L11.H2 & C & $-8.5$ & $[-10.5,-6.8]$ \\
L12.H7 & C & $-29.0$ & $[-35.2,-23.4]$ & L12.H0 & C & $-7.8$ & $[-12.4,-3.1]$ \\
 & & &  & L13.H2 & C & $-7.0$ & $[-9.8,-4.4]$ \\
\bottomrule
\end{tabular}
\end{table}

\clearpage
\section{Group effects, accuracy, and evaluation stability}
\label{app:groupcanc}
\label{app:group_intervals}
The complete group effects and task accuracies appear in
Tables~\ref{tab:group_intervals} and \ref{tab:accuracy}.
Zero and mean ablation use the same $500$ evaluation prompts within
each setting. The intervals come from paired-prompt bootstraps.

\begin{table}[!ht]
\centering
\footnotesize
\setlength{\tabcolsep}{5pt}
\renewcommand{\arraystretch}{1.06}
\caption{\textbf{Main group effects with uncertainty.} Readout shifts in nats and paired-bootstrap $95\%$ confidence intervals.}
\label{tab:group_intervals}
\begin{tabular}{lccc}
\toprule
Setting & $\delta_{\mathcal W}$ [CI] & $\delta_{\mathcal C}$ [CI] & $\delta_{\mathcal F}$ [CI] \\
\midrule
\multicolumn{4}{l}{\textit{Zero ablation}} \\
hier-410M & $-0.248\;[-0.323,-0.175]$ & $+0.288\;[+0.200,+0.376]$ & $+0.002\;[-0.068,+0.071]$ \\
hier-1B & $-0.558\;[-0.743,-0.369]$ & $+0.160\;[+0.119,+0.201]$ & $-0.430\;[-0.595,-0.264]$ \\
hier-1.4B & $-0.321\;[-0.413,-0.227]$ & $+0.150\;[+0.094,+0.206]$ & $-0.128\;[-0.195,-0.059]$ \\
mod-410M & $-0.391\;[-0.531,-0.252]$ & $+0.165\;[+0.119,+0.212]$ & $-0.144\;[-0.262,-0.025]$ \\
mod-1B & $-0.993\;[-1.211,-0.766]$ & $+0.206\;[+0.129,+0.283]$ & $-0.784\;[-0.966,-0.599]$ \\
mod-1.4B & $-0.486\;[-0.618,-0.354]$ & $+0.130\;[+0.089,+0.170]$ & $-0.375\;[-0.467,-0.282]$ \\
\midrule
\multicolumn{4}{l}{\textit{Mean ablation}} \\
hier-410M & $-0.343\;[-0.456,-0.229]$ & $+0.290\;[+0.195,+0.385]$ & $-0.172\;[-0.266,-0.077]$ \\
hier-1B & $-0.444\;[-0.615,-0.273]$ & $+0.276\;[+0.214,+0.337]$ & $-0.337\;[-0.490,-0.183]$ \\
hier-1.4B & $-0.370\;[-0.500,-0.239]$ & $+0.224\;[+0.154,+0.295]$ & $-0.236\;[-0.341,-0.129]$ \\
mod-410M & $-0.312\;[-0.457,-0.167]$ & $+0.125\;[+0.070,+0.181]$ & $-0.232\;[-0.361,-0.102]$ \\
mod-1B & $-0.400\;[-0.552,-0.246]$ & $+0.263\;[+0.163,+0.363]$ & $-0.254\;[-0.379,-0.128]$ \\
mod-1.4B & $-0.348\;[-0.486,-0.210]$ & $+0.093\;[+0.047,+0.139]$ & $-0.255\;[-0.368,-0.144]$ \\
\bottomrule
\end{tabular}
\end{table}

\subsection{Task accuracy}
\label{app:accuracy}
Table~\ref{tab:accuracy} reports accuracy under the same group
interventions as Table~\ref{tab:group_intervals}.

\begin{table}[!ht]
\centering
\caption{\textbf{Observed task accuracy (\%).} Baseline accuracy and accuracy
under writer ($\mathcal W$), canceller ($\mathcal C$), and joint
($\mathcal F$) interventions. Each setting uses $500$ prompts.}
\label{tab:accuracy}
\small
\setlength{\tabcolsep}{4pt}
\begin{tabular}{lrrrr}
\toprule
Setting & Baseline & $\mathcal W$ & $\mathcal C$ & $\mathcal F$ \\
\midrule
\multicolumn{5}{l}{\textit{Zero ablation}} \\
hier-410M & $54.6$ & $49.4$ & $61.4$ & $57.6$ \\
hier-1B & $57.6$ & $41.6$ & $60.0$ & $45.6$ \\
hier-1.4B & $60.8$ & $53.6$ & $65.4$ & $60.4$ \\
mod-410M & $55.6$ & $54.8$ & $59.0$ & $64.0$ \\
mod-1B & $54.6$ & $31.2$ & $58.2$ & $38.6$ \\
mod-1.4B & $56.2$ & $55.6$ & $59.8$ & $50.4$ \\
\midrule
\multicolumn{5}{l}{\textit{Mean ablation}} \\
hier-410M & $54.6$ & $49.2$ & $62.0$ & $57.6$ \\
hier-1B & $57.6$ & $43.0$ & $61.8$ & $59.2$ \\
hier-1.4B & $60.8$ & $56.2$ & $62.8$ & $63.4$ \\
mod-410M & $55.6$ & $59.6$ & $58.4$ & $66.6$ \\
mod-1B & $54.6$ & $63.4$ & $56.4$ & $67.4$ \\
mod-1.4B & $56.2$ & $63.0$ & $56.0$ & $64.4$ \\
\bottomrule
\end{tabular}
\end{table}

\clearpage
\subsection{Evaluation stability}
\label{app:dla}
\label{app:split_detail}
Each setting has five evaluation splits, indexed by seeds $0$--$4$.
Each row of Table~\ref{tab:split_all} uses $500$ prompts in that split's
evaluation half. Group membership stays fixed from paired validation,
so these results measure the stability of the same groups under
evaluation resampling.

\begin{table}[!ht]
\centering
\footnotesize
\setlength{\tabcolsep}{5pt}
\renewcommand{\arraystretch}{1.06}
\caption{\textbf{All evaluation-split effects.} Group-ablation shifts in nats with $95\%$ confidence intervals. The three interventions share the evaluation prompts in each row.}
\label{tab:split_all}
\begin{tabular}{llccc}
\toprule
Task & Split seed & $\delta_{\mathcal W}$ [CI] & $\delta_{\mathcal C}$ [CI] & $\delta_{\mathcal F}$ [CI] \\
\midrule
\multicolumn{5}{c}{\textbf{Pythia-410M}} \\
Hierarchical & 0 & $-0.298\;[-0.378,-0.215]$ & $+0.274\;[+0.185,+0.364]$ & $-0.057\;[-0.123,+0.010]$ \\
Hierarchical & 1 & $-0.306\;[-0.381,-0.231]$ & $+0.296\;[+0.208,+0.384]$ & $-0.052\;[-0.121,+0.019]$ \\
Hierarchical & 2 & $-0.309\;[-0.381,-0.235]$ & $+0.281\;[+0.193,+0.367]$ & $-0.058\;[-0.127,+0.012]$ \\
Hierarchical & 3 & $-0.320\;[-0.400,-0.239]$ & $+0.276\;[+0.188,+0.363]$ & $-0.067\;[-0.133,+0.000]$ \\
Hierarchical & 4 & $-0.280\;[-0.355,-0.204]$ & $+0.236\;[+0.145,+0.323]$ & $-0.073\;[-0.141,-0.006]$ \\
Modular & 0 & $-0.426\;[-0.566,-0.287]$ & $+0.153\;[+0.102,+0.206]$ & $-0.189\;[-0.311,-0.068]$ \\
Modular & 1 & $-0.384\;[-0.522,-0.243]$ & $+0.151\;[+0.097,+0.201]$ & $-0.140\;[-0.257,-0.022]$ \\
Modular & 2 & $-0.545\;[-0.686,-0.403]$ & $+0.171\;[+0.123,+0.217]$ & $-0.284\;[-0.404,-0.162]$ \\
Modular & 3 & $-0.495\;[-0.640,-0.351]$ & $+0.152\;[+0.101,+0.201]$ & $-0.259\;[-0.383,-0.137]$ \\
Modular & 4 & $-0.345\;[-0.485,-0.201]$ & $+0.160\;[+0.109,+0.209]$ & $-0.097\;[-0.217,+0.028]$ \\
\midrule
\multicolumn{5}{c}{\textbf{Pythia-1B}} \\
Hierarchical & 0 & $-0.650\;[-0.851,-0.440]$ & $+0.147\;[+0.105,+0.190]$ & $-0.548\;[-0.727,-0.363]$ \\
Hierarchical & 1 & $-0.656\;[-0.842,-0.471]$ & $+0.182\;[+0.140,+0.223]$ & $-0.525\;[-0.691,-0.359]$ \\
Hierarchical & 2 & $-0.650\;[-0.837,-0.460]$ & $+0.176\;[+0.134,+0.216]$ & $-0.514\;[-0.684,-0.345]$ \\
Hierarchical & 3 & $-0.672\;[-0.868,-0.475]$ & $+0.156\;[+0.113,+0.200]$ & $-0.551\;[-0.724,-0.378]$ \\
Hierarchical & 4 & $-0.625\;[-0.817,-0.434]$ & $+0.141\;[+0.098,+0.183]$ & $-0.509\;[-0.683,-0.342]$ \\
Modular & 0 & $-1.057\;[-1.268,-0.851]$ & $+0.181\;[+0.099,+0.259]$ & $-0.850\;[-1.030,-0.680]$ \\
Modular & 1 & $-0.982\;[-1.196,-0.765]$ & $+0.201\;[+0.120,+0.278]$ & $-0.787\;[-0.963,-0.607]$ \\
Modular & 2 & $-1.133\;[-1.343,-0.918]$ & $+0.224\;[+0.147,+0.301]$ & $-0.863\;[-1.038,-0.686]$ \\
Modular & 3 & $-1.076\;[-1.296,-0.853]$ & $+0.207\;[+0.128,+0.285]$ & $-0.855\;[-1.038,-0.668]$ \\
Modular & 4 & $-0.913\;[-1.128,-0.693]$ & $+0.204\;[+0.124,+0.283]$ & $-0.717\;[-0.900,-0.533]$ \\
\midrule
\multicolumn{5}{c}{\textbf{Pythia-1.4B}} \\
Hierarchical & 0 & $-0.377\;[-0.480,-0.270]$ & $+0.131\;[+0.076,+0.185]$ & $-0.170\;[-0.238,-0.101]$ \\
Hierarchical & 1 & $-0.376\;[-0.469,-0.281]$ & $+0.120\;[+0.065,+0.176]$ & $-0.195\;[-0.259,-0.129]$ \\
Hierarchical & 2 & $-0.401\;[-0.496,-0.304]$ & $+0.142\;[+0.087,+0.195]$ & $-0.184\;[-0.249,-0.117]$ \\
Hierarchical & 3 & $-0.363\;[-0.464,-0.262]$ & $+0.128\;[+0.076,+0.181]$ & $-0.169\;[-0.236,-0.103]$ \\
Hierarchical & 4 & $-0.343\;[-0.440,-0.247]$ & $+0.109\;[+0.053,+0.165]$ & $-0.177\;[-0.244,-0.111]$ \\
Modular & 0 & $-0.507\;[-0.633,-0.384]$ & $+0.108\;[+0.065,+0.149]$ & $-0.414\;[-0.506,-0.326]$ \\
Modular & 1 & $-0.477\;[-0.605,-0.347]$ & $+0.118\;[+0.073,+0.161]$ & $-0.392\;[-0.485,-0.297]$ \\
Modular & 2 & $-0.582\;[-0.710,-0.451]$ & $+0.145\;[+0.105,+0.185]$ & $-0.419\;[-0.510,-0.327]$ \\
Modular & 3 & $-0.557\;[-0.688,-0.424]$ & $+0.127\;[+0.085,+0.168]$ & $-0.426\;[-0.519,-0.331]$ \\
Modular & 4 & $-0.434\;[-0.561,-0.305]$ & $+0.130\;[+0.088,+0.170]$ & $-0.327\;[-0.421,-0.232]$ \\
\bottomrule
\end{tabular}
\end{table}

\clearpage
\section{Source contributions and value interventions}
\label{app:qksource}
\label{app:contentvssink}
For seven canceller assignments, BOS or format positions dominate
the source attribution. Six belong to three heads that appear in
both rule tasks, L14.H3 at $410$M, L10.H2 at $1$B, and L11.H2 at
$1.4$B. Almost all of L11.H4's negative DLA comes from
demonstration-label positions in both $410$M settings.

Table~\ref{tab:valueshuffle} reports source-value shuffling for the
dominant canceller in each setting. L11.H4 uses $200$ prompts and
three shuffle seeds. The remaining settings use $500$ prompts and
five shuffle seeds.

\begin{table}[!ht]
\centering
\caption{\textbf{Source-value shuffling.} Fraction of the dominant
canceller's negative DLA removed by permuting values at fixed
attention weights.}
\label{tab:valueshuffle}
\small
\begin{tabular}{llr}
\toprule
Setting & Head & Fraction removed \\
\midrule
hier-410M & L11.H4 & $0.817$ \\
hier-1B & L13.H1 & $0.777$ \\
hier-1.4B & L12.H7 & $0.827$ \\
mod-410M & L11.H4 & $0.523$ \\
mod-1B & L13.H1 & $0.833$ \\
mod-1.4B & L10.H7 & $0.770$ \\
\bottomrule
\end{tabular}
\end{table}

For L11.H4, hierarchical-task DLA changes from $-0.216$ to
$-0.040$, a shift of $+0.176$ (CI $[+0.080,+0.279]$).
Modular-task DLA changes from $-0.215$ to $-0.103$, a shift of
$+0.113$ (CI $[+0.015,+0.206]$).

\section{Transfer across tasks}
\label{app:rule_transfer_detail}
\subsection{Transfer between rule families}
For each model size, the groups identified on one rule family are
applied to the other with membership fixed. The target keeps the
four-shot format and binary answer labels while changing the
input--output rule.

\begin{table}[!ht]
\centering
\footnotesize
\setlength{\tabcolsep}{5pt}
\renewcommand{\arraystretch}{1.06}
\caption{\textbf{Transfer between rule families.} Group-ablation shifts in the target task's correct-label readout, with $95\%$ confidence intervals. Each comparison uses $500$ target-task prompts.}
\label{tab:rule_transfer_detail}
\begin{tabular}{lllccc}
\toprule
Source & Target & Model & $\delta_{\mathcal W}$ [CI] & $\delta_{\mathcal C}$ [CI] & $\delta_{\mathcal F}$ [CI] \\
\midrule
Hier. & Mod. & 410M & $-0.258\;[-0.338,-0.176]$ & $+0.327\;[+0.244,+0.410]$ & $+0.062\;[-0.007,+0.132]$ \\
Hier. & Mod. & 1B & $-0.789\;[-0.998,-0.574]$ & $+0.097\;[+0.060,+0.132]$ & $-0.769\;[-0.956,-0.576]$ \\
Hier. & Mod. & 1.4B & $-0.368\;[-0.472,-0.262]$ & $+0.171\;[+0.121,+0.222]$ & $-0.084\;[-0.150,-0.017]$ \\
Mod. & Hier. & 410M & $-0.297\;[-0.430,-0.163]$ & $+0.146\;[+0.095,+0.197]$ & $-0.098\;[-0.219,+0.022]$ \\
Mod. & Hier. & 1B & $-0.573\;[-0.762,-0.384]$ & $+0.226\;[+0.140,+0.309]$ & $-0.378\;[-0.544,-0.211]$ \\
Mod. & Hier. & 1.4B & $-0.317\;[-0.429,-0.204]$ & $+0.112\;[+0.068,+0.155]$ & $-0.238\;[-0.324,-0.153]$ \\
\bottomrule
\end{tabular}
\end{table}

\subsection{Transfer to vocabulary tasks}
\label{app:vocab}
The antonym and country-capital datasets contain $60$ and $50$
word pairs, respectively. Each prompt has four demonstrations in
\texttt{word:answer} format. Effects are evaluated on $100$ prompts
per source--target comparison, using the source-task groups.

\begin{table}[!ht]
\centering
\caption{\textbf{Transfer to vocabulary tasks.} Correct-answer logit shifts
in nats after zero-ablating the source-task groups. Positive shifts
indicate that ablation raises the correct-answer logit.}
\label{tab:vocabtransfer}
\small
\setlength{\tabcolsep}{3pt}
\begin{tabular}{llrrr}
\toprule
Source & Target & $\delta_{\mathcal W}$ & $\delta_{\mathcal C}$ & $\delta_{\mathcal F}$ \\
\midrule
hier-410M & antonym         & $-0.48$ & $-1.56$ & $-2.00$ \\
hier-410M & country-capital & $-0.52$ & $+0.14$ & $-0.37$ \\
mod-410M  & antonym         & $-0.33$ & $+0.03$ & $-0.24$ \\
mod-410M  & country-capital & $-0.59$ & $+0.23$ & $-0.36$ \\
\bottomrule
\end{tabular}
\end{table}

At single-head resolution, L11.H4's zero-ablation shift is
$-0.069$ (CI $[-0.109,-0.028]$) on antonyms and $-0.005$
(CI $[-0.033,+0.025]$) on country-capital.

\section{Magnitude ranking}
\label{app:fv_overlap}
Table~\ref{tab:fv_overlap_full} gives MA-$20$ coverage by setting.
The pooled recovery percentages combine the three model sizes
within each rule family, using the total number of validated
writers or cancellers as the denominator.

\begin{table}[!ht]
\centering
\caption{\textbf{Coverage of signed roles by magnitude ranking.}
MA-$20$ contains the heads with the largest absolute mean-ablation
effects on each task. The final two columns count its overlap with
the validated writer and canceller groups. Jaccard overlap compares
MA-$20$ with their union $\mathcal F$.}
\label{tab:fv_overlap_full}
\small
\setlength{\tabcolsep}{3.2pt}
\begin{tabular}{lccccc}
\toprule
Setting & $|\mathcal{W}|$ & $|\mathcal{C}|$ & Jaccard & W$\in$MA20 & C$\in$MA20 \\
\midrule
hier-410M & $8$  & $6$ & $0.06$ & $0$  & $\mathbf{2}$ \\
hier-1B   & $9$  & $4$ & $0.18$ & $1$  & $\mathbf{4}$ \\
hier-1.4B & $10$ & $4$ & $0.10$ & $0$  & $\mathbf{3}$ \\
mod-410M  & $8$  & $3$ & $0.19$ & $\mathbf{5}$  & $0$ \\
mod-1B    & $8$  & $6$ & $0.17$ & $\mathbf{5}$  & $0$ \\
mod-1.4B  & $11$ & $4$ & $0.25$ & $\mathbf{6}$  & $1$ \\
\midrule
\textbf{hier total} & $27$ & $14$ & $\cdot$ & $\mathbf{1\,(4\%)}$ & $\mathbf{9\,(64\%)}$ \\
\textbf{mod total}  & $27$ & $13$ & $\cdot$ & $\mathbf{16\,(59\%)}$ & $\mathbf{1\,(8\%)}$ \\
\bottomrule
\end{tabular}
\end{table}

\end{document}